%% file: main.tex
\documentclass[review]{elsarticle}
\makeatletter
\def\set@curr@file#1{%
\begingroup
  \escapechar\m@ne
  \xdef\@curr@file{\expandafter\string\csname #1\endcsname}%
\endgroup
}
\def\quote@name#1{"\quote@@name#1\@gobble""}
\def\quote@@name#1"{#1\quote@@name}
\def\unquote@name#1{\quote@@name#1\@gobble"}
\makeatother

\usepackage{lineno}

\usepackage[utf8x]{inputenc}
\usepackage[T1]{fontenc}
\usepackage{natbib}
\biboptions{authoryear}
\usepackage{graphicx}
\usepackage[english]{babel}
\usepackage{amssymb}
\usepackage[cmex10]{amsmath}
\usepackage{algorithm}
\usepackage{algpseudocode}
\usepackage{array}
\usepackage[tight,footnotesize]{subfigure}
\usepackage{fixltx2e}
\usepackage{microtype}
\usepackage[binary-units=true]{siunitx}
\usepackage{psfrag}
\usepackage{pst-sigsys}
\usepackage{pstricks}
\usepackage{pgfplots}
\usepackage{pgfkeys}
\usepackage{booktabs}
\usepackage{ae}
\usepackage{url}
\usepackage[nolist,nohyperlinks]{acronym}
\usepackage{siunitx}
\input{tikz_pck}
\newlength\figureheight
\newlength\figurewidth

\usepackage{subfigure}
\usepackage{graphicx, amssymb, amsmath, amsthm}
\usepackage[colorlinks,citecolor=blue,linkcolor=blue]{hyperref}
\usepackage{nicefrac}
\usepackage[normalem]{ulem}
\usepackage{color}
\usepackage{multirow}
\usepackage{hyperref}
\usepackage{setspace}
\usepackage{titlesec}
\usepackage{siunitx}
\usepackage{enumitem}
\usepackage{algorithm}
\usepackage{algpseudocode}

\theoremstyle{definition}

\newcommand{\norm}[1]{\left\Vert#1\right\Vert}

\newcommand{\ui}[2]{#1_{\rm #2}}
\newcommand{\Ts}{\ui{T}{s}}
\newcommand{\nx}{\ui{n}{x}}
\renewcommand{\nu}{\ui{n}{u}}

\journal{Acta Chimica Slovaca}

\definecolor{amber}{rgb}{1.0, 0.49, 0.0}
\definecolor{cadmiumorange}{rgb}{0.93, 0.53, 0.18}
\definecolor{darkorange}{rgb}{1.0, 0.55, 0.0}
 \definecolor{royalazure}{rgb}{0.0, 0.22, 0.66}
\definecolor{blue(pigment)}{rgb}{0.2, 0.2, 0.6}
\definecolor{bakermillerpink}{rgb}{1.0, 0.57, 0.69}

\begin{document}

\begin{frontmatter}

\title{Neural Network Based Explicit MPC for Chemical Reactor Control}

\author{Karol Ki\v s}
\author{Martin Klau\v co}

\address{Slovak University of Technology in Bratislava,\\
Radlinsk\'eho 9, SK-812 37 Bratislava, Slovak Republic \\
{\tt\small\{karol.kis,martin.klauco\}@stuba.sk}, +421 259 325 
345}

\begin{abstract}
	In this paper, we show the implementation of deep neural networks applied in process control. In our approach, we based the training of the neural network on model predictive control. Model predictive control is popular for its ability to be tuned by the weighting matrices and by the fact that it respects the constraints. We present the neural network that can approximate the behavior of the MPC in the way of mimicking the control input trajectory while the constraints on states and control input remain unimpaired of the value of the weighting matrices. This approach is demonstrated in a simulation case study involving a continuous stirred tank reactor, where multi-component chemical reaction takes place.
			
\end{abstract}
\end{frontmatter}

\section{Introduction}
Chemical reactors play an essential role in the chemical and petrochemical industry. Their vast presence in the industrial world makes the control synthesis very attractive for researchers. Throughout recent years, several new control approaches arise mainly from optimal control theory, as discussed by~\cite{smets:2004:jpc:bioreact,pourdehi:2018:jpc:react,bakosova:2012:tfce:rmpc}. The optimal control theory proved to be very promising, due to its natural ability to cope with technological constraints and following a performance criterion, which defines the overall economy of the production. 

Designers of control strategies for chemical reactors must cope with several obstacles, mainly, the natural instability of the process, e.g., exothermic reactors, and keeping the process variables in their designed steady state. Alongside which the controller must be able to decrease the energy consumption and increase the quality of the product. All control objectives can be incorporated in an optimal control problem (OCP)~\citep{bakosova:2014:acs,singh:2010:ifac}. A model predictive control (MPC) technique is often used in such control tasks~\cite{Prasath:2010}, since its construction is straight-forward~\citep{maciejowski:book:2002}. 

The core concept of the MPC is to predict the future evolution of the controlled variables based on current measurements. Then, with respect to a quality criterion (usually energy consumption), the MPC optimizes the values of the manipulated variables, such that the criterion is minimized~\citep{klauco:2019:book}.

%

Even though it seems that the MPC is one of the best controllers, it has several drawbacks. Since it is an optimization-based controller, it requires a repeated solution to an optimal control problem. Such an arrangement is virtually impossible to implement in the industry or on the computers responsible for the operation of the chemical reactors.

The traditional way of coping with this limitation is to consider explicit model predictive control (EMPC). The explicit MPC is an analytical solution to the optimal control problem~\citep{bemporad:aut:2002}. The control law given by the explicit solution is in the form of piecewise affine function (PWA) \citep{borrelli:2017:book}. Such control law can be easily evaluated at any given time, without the need for involving the optimization procedure. In other words, it allows us to replace the optimization solver with function evaluation. The EMPC, unfortunately, can be constructed only for small-sized systems with short prediction horizons. Such a property is especially a significant limitation in process industries, where prediction horizons are long.

This paper proposes an alternative to the explicit controller that is based on neural networks. The neural network is a powerful mathematical concept, that is capable of approximating an arbitrary continuous function~\citep{hornik:1991:nn}. Here, we propose to approximate the explicit control law given by the full-fidelity MPC. Since we do not consider the traditional explicit model predictive control, which results in a PWA function, we are not limited by the length of the prediction horizon, or by the size of the controlled system. Similar work has been done by~\cite{SergioMPC} or by~\cite{klaucoMimicking}. In this paper, however, we focus on application in chemical technology, mainly the control of the multicomponent chemical reaction. 

\section{Theoretical}
Firstly, we present the optimal control problem (OCP) that stands for the model predictive controller. The third part of this section is devoted to the artificial neural network, which will substitute the model predictive controller.

\subsection{Model Predictive Control}
The standard formulation of the model predictive controller utilizes a linear time-invariant model that captures the dynamics of the controlled process. Specifically, we consider a discrete-time dynamics, given as
\begin{subequations}
    \label{eq:ss}
    \begin{align}
        x(t+\Ts) &= Ax(t) + Bu(t),\\
        y(t) &= Cx(t) + Du(t),
    \end{align}
\end{subequations}
where the variables $x \in \mathbb{R}^{\nx}$ stands for process state variables, vector $u \in\mathbb{R}^{\nu}$ represents the manipulated variables and $y \in\mathbb{R}^{\nu}$ depicts the process variable. Matrices $A\in \mathbb{R}^{n_x \times n_x}$, $B\in \mathbb{R}^{n_x \times n_u}$, $C\in \mathbb{R}^{n_y \times n_x}$, and $D\in \mathbb{R}^{n_y \times     n_u}$ are obtained from a dynamical model representing an actual controlled process using first order Taylor expansion. The discrete time linear model~\eqref{eq:ss} is discretized with sampling period of $\Ts$.

The model predictive control is then constructed as follows
\begin{subequations}
	\label{eq:mpc}
	\begin{align}
	\min_{u_0,\ldots, u_{N-1}} \; & \; \sum _{k = 0} ^{N-1} y_k^T Q y_k + 
	\sum _{k = 0} ^{N-1} u_k^T R u_k, 
	\label{eq:mpc:du:cost_r}\\
	\text{s.t.} \; & \; x_{k+1} = Ax_k + Bu_k \quad k = 0, \ldots, N-1, 
	\label{eq:mpc:du:prediction_x_r} \\
	& \; y_{k} = Cx_k + Du_k \quad k = 0, \ldots, N-1, 
	\label{eq:mpc:du:prediction_y_r} \\
	& \; \ui{y}{min} \leq x_k \leq \ui{y}{max}  \quad k = 1, \ldots, N-1,
	\label{eq:mpc:du:cst_x_r} \\
	& \; \ui{u}{min} \leq u_k \leq \ui{u}{max} \quad k = 1, \ldots, N-1,
	\label{eq:mpc:du:cst_u_r} \\
	& \; x_0 = x(t), \label{eq:mpc:x0}
	\end{align}
\end{subequations}
where $N$ denotes the prediction horizon. Next, the cost function~\eqref{eq:mpc:du:cost_r} is in the form of convex quadratic function, and with positive definite tuning factors $Q \in \mathbb{R}^{n_x \times n_x}$ and $R \in \mathbb{R}^{n_u \times n_u}$. The objective function is posed such that, the controlled variables are driven towards the steady-state. Moreover, we embed the technological constraints as min-max limits, on controlled variable and on manipulated variables, as in~\eqref{eq:mpc:du:cst_x_r} and~\eqref{eq:mpc:du:cst_u_r}, respectively. The optimization problem is initialized with measurement $x(t)$. The MPC is formulated as quadratic optimization problem with linear constraints. The optimal solution to the MPC in~\eqref{eq:mpc} yields an optimal sequence of manipulated variables $[u_0^\star, \ldots, u_{N-1}^\star]^T$. Since the OCP is a convex optimization problem its solution is a global minimum.

The process is controlled by the model predictive controller using an algorithm called receding horizon policy, presented and proven by~\cite{mayne:aut:2000} and is given as:
\begin{enumerate}
	\item Measure system process variables $x(t)$ (e.g. temperature, or concentration).
	\item Initialize the MPC in~\eqref{eq:mpc} with $x(t)$.
	\item Solve the quadratic optimization problem.
	\item Apply manipulated variable $u_0$.
	\item After $\Ts$ continue from step 1.
\end{enumerate}
We refer to this algorithm as to a closed-loop implementation of the MPC. The bottleneck of the algorithm is the step No. $3$, where one must solve an optimization problem. In average industrial application, this is an impossible tasks, since there are no machines present, that are capable of solving complex mathematical problem, and second, the solver impose additional costs. 

In the next section, we present how such an algorithm can be replaced with neural network, in order to substitute the complex procedure of mathematical optimization performed every sampling instant.

\subsection{Artificial Neural Networks}
\begin{figure}[ht]
	\centering
	\includegraphics[width=0.8\linewidth]{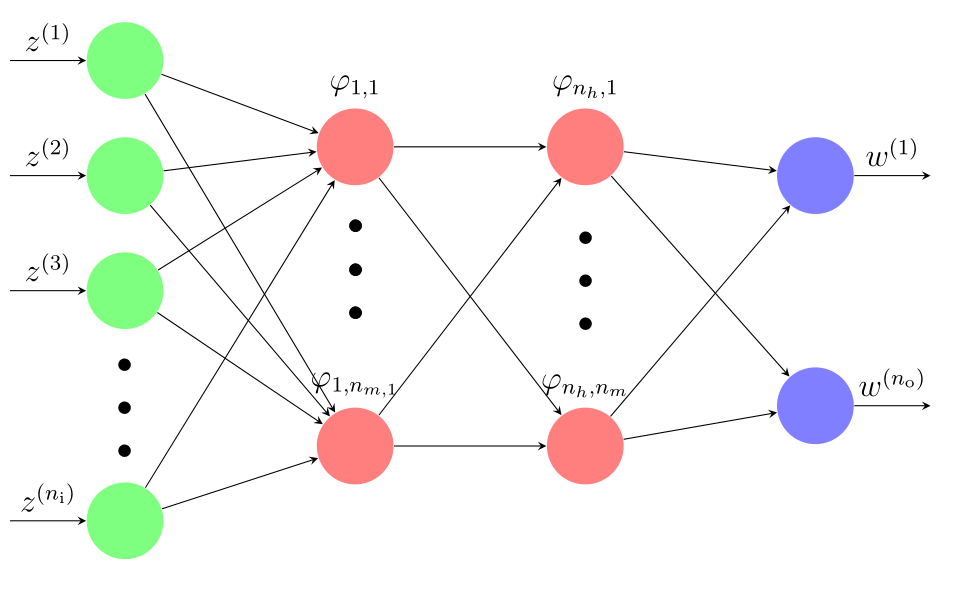}
	\caption{Example of the structure of the neural network. Green points represent an input layer, red dots depicts the hidden layers, and the output layer is colored by blue.}
	\label{fig:nn}
\end{figure}

The neural network is a mathematical function, that maps inputs $z \in \mathbb{R}^{n_z} \to w \in \mathbb{R}^{n_w}$, via interconnected monotone functions. Structure of the neural network is visualized in the figure~\ref{fig:nn}. On the figure, each green and red dot represents the monotone function, also called as activation functions, and they are given as
\begin{equation}
	\label{eq:nn_act}
	\varphi(\alpha, z) = \frac{2}{1 + {\rm e}^{\alpha z}} - 1,
\end{equation}
where by the $z$ we denote an aggregated input to each node, while the $\alpha$ is a tuning parameter of the activation function. The blue dots stands for an linear output layer.

The neural network is capable of approximating, with a very high confidence number, any arbitrary continuous functions as discussed by~\cite{hornik:1991:nn}. Another significant advantage is the explicit nature of the neural net. Once, we will construct a neural net of suitable properties, and it can be evaluated on moderate hardware, hence no need for optimization solver. Another main advantage is that the neural net is in no way limited by the length of the prediction horizon or by the size of the system, compared to the explicit model predictive control. The structure of the NN-based controller with $\alpha$ values determines the goodness of the approximation of the PWA control law. The procedure of getting weights is called the training of the neural network, and it is performed for a fixed structure of the neural net, hence only the weights are calculated.

To give the reader an illustration, how the weights of activation functions are calculated, consider
\begin{equation}
	\label{eq:nn_min}
        \min _{\alpha} \; \sum_{k = 1}^{\ui{n}{k}} \Big(
            \tilde{w}_k - \varphi(\alpha, z_k)
        \Big)^2.
\end{equation}
The optimization problem~\eqref{eq:nn_min} is a sum-of-squares data fitting problem. Solution to~\eqref{eq:nn_min} gives optimal values of the weight $\alpha$, based on minimizing the squared distance between a target value $\tilde{w}$ and evaluated activation function $\varphi(\alpha, z_k)$, for a given neural network input. Naturally, to increase the goodness of the $\alpha$ value, a sufficient number of data points must be included in the minimization procedure. Here, the $n_k$ denotes the number of those data points.

Since the neural network is substituting the controller, the input to the training of the NN-based controller are measurements of the process variables $x(t)$, and the training targets are the control inputs, i.e., the manipulated variable $u(t)$. To provide a suitable basis for the training procedure, we construct an initial training set, given as 
\begin{equation}
    \mathcal{X} = \begin{bmatrix}
        x_{min}, \ldots, x_{max} \\
    \end{bmatrix},
    \label{eq:unidistrpoints}
\end{equation}
where we equidistantly grid the limits on the process variables with $n_k$ points. Subsequently, for each data point in the $\mathcal{X}$, we calculate a corresponding control action. Thus, we obtain set $\mathcal{U}$, with $n_k$ different manipulated variables. The corresponding control actions are obtained by solving the MPC problem~\eqref{eq:mpc}, initialized with $x_0$ from~\eqref{eq:mpc:x0} equal to one of the points from $\mathcal{X}$. Sets $\mathcal{X}$ and $\mathcal{U}$ together form a learning data set, from which the minimization problem~\eqref{eq:nn_min} is constructed. Note, that the problem~\eqref{eq:nn_min} is presented only for one node in the hidden layer, the overall training procedure consists of an aggregated minimization problem, where all nodes are included. 

Numerically, the optimal control actions are obtained using the GUROBI solver, while the training procedure is performed with the \emph{Deep Learning Toolbox} in MATLAB. Furthermore, the model predictive controller is formulated using YALMIP toolbox~\citep{yalmip:paper}.

\section{Experimental}
The theory presented in the Theoretical part of the paper is applied on a case study involving the control of a multi-component chemical reactor. Specifically, we consider benchmark chemical reaction 
\begin{equation}
	A \leftrightarrows 2C \rightarrow B,
\end{equation}
with dynamical behavior given by three differential equations~\citep{fissore2008robust,PBAC2018}, that reads to
\begin{subequations}
	\label{eq:cstr_nonlin}
	\begin{align}
	\dot{c}_{\text{A}} & = -k_1 c_{\text{A}} + \frac{F}{V}(c_{A,
		\text{feed}} - c_{\text{A}}) + k_2 c_{\text{C}}^2,\\
	\dot{c}_{\text{B}} & = -\frac{F}{V} c_{\text{B}} + k_3
	c_{\text{C}}^2,\\
	\dot{c}_{\text{C}} & = k_1 c_{\text{A}} - \frac{F}{V} c_{\text{C}}
	- (k_2 + k_3) c_{\text{C}}^2 + q_{in}.
	\end{align}
\end{subequations}
The process variables are the concentrations $x = [{c}_{\text{A}},\;{c}_{\text{B}},\;{c}_{\text{C}}]^\intercal$, while the manipulated variables $q_{in}$ stands for molar feed of the component $C$. Specific parameters of the benchmark model of the chemical reactor are reported in the table~\ref{tab:params}. The objective of the controller is to keep the concentration ${c}_{\text{B}}$ at the steady state level, which represents optimal conditions of the reactor operation, as introduced by~\cite{fissore2008robust}. 

\begin{table}[ht]
	\begin{center}
		\caption{Table of model parameters.}
		\label{tab:params}		
		\begin{tabular}{ccc}
			\toprule
			\textbf{Variable} & \textbf{Value} & \textbf{Unit} \\
			\midrule
				$k_1$ & 1 & \SI{}{\cubic\metre\per\mol\per\second} \\ 
				$k_2$ & 3 & \SI{}{\cubic\metre\per\mol\per\second} \\ 
				$k_3$ & 5 & \SI{}{\cubic\metre\per\mol\per\second} \\ 
				$F$ & 3 & \SI{}{\cubic\metre\per\second} \\ 
				$V$ & 3 & \SI{}{\cubic\metre}\\ 
				$c_{\text{A},\text{feed}}$ & 2 & \SI{}{\mol\per\cubic\metre} \\ 
			\bottomrule
		\end{tabular}
	\end{center}
\end{table}

Concretely, the optimal operation of the chemical reactor is given by a set of steady-state values of individual process variables, and they are given as
\begin{equation}
	c_{\text{A}, \text{S}} = \SI{2.18}{\mol\per\cubic\metre},
	c_{\text{B}, \text{S}} = \SI{3.93}{\mol\per\cubic\metre},
	c_{\text{C}, \text{S}} = \SI{0.87}{\mol\per\cubic\metre},
\end{equation}
while the steady-state manipulated variables was set to $q_{in,\text{S}} = \SI{5}{\mol\per\second}$. 

Next, the model predictive controller was constructed with the linearized version of the dynamical mathematical model, which was sampled with $T_s = \SI{0.1}{\second}$. The matrices of the discretized state-space model take the form
\begin{subequations}
	\label{eq:ss_matrices}
	\begin{align}
	\text{A} &= \begin{bmatrix}
	0.83     &    0  &  0.24\\
	0.03  &  0.90  &  0.43\\
	0.05    &     0  &  0.23
	\end{bmatrix},\quad
	\text{B} = \begin{bmatrix}
	0.02\\
	0.03\\
	0.05
	\end{bmatrix},\\
	C &= \begin{bmatrix}
		0 & 1& 0
	\end{bmatrix}.
	\end{align}
\end{subequations}
Next, the MPC was set up with prediction horizon $N = 50$, while the tuning factors were set to $Q = 10$ and $R = 0.15$. The constraints were again reproduced from the benchmark model, where $0 \leq c_{\text{A}} \leq 10$, $0 \leq c_{\text{B}} \leq 14$, and $0 \leq c_{\text{C}} \leq 1.1$. The molar feed-flow of the $q_{in}$ is constrained to interval $[0,\;10]\SI{}{\mole\per\second}$. 

Next, the learning set for the neural network was constructed. Each interval for the process variables was split into $10000$ samples, for which we obtained the corresponding optimal value of the manipulated variable. Such a training set was then fed into the \emph{Deep Learning Toolbox}, particularly, the \texttt{fitnet} command, which trained the neural network. The structure of the NN-based controller consists of $3$ nodes in the input layer (due to $3$ individual measurements of the process variable) and then from $4$ hidden layers. Each hidden layer consists of $4$ nodes, where each node has the form of the action function as in~\eqref{eq:nn_act}. The final output layer has a linear structure and consists of $1$ node. Recall that the output from the neural network is the manipulated variable. The training is done offline, and it took \SI{80}{\second} on a personal computer with Core i7, 16GB of RAM, and Matlab R2019a. Resulting neural network controller takes less than $7$kB of memory and can be evaluated in milli-second range on ARM processors. Such a characteristic is in stark contrast to the optimization procedure, which requires the MPC strategy.
\begin{figure}[ht]
	\vspace{-40pt}
	\centering
	\subfigure{\includegraphics[height = 0.17\textheight]{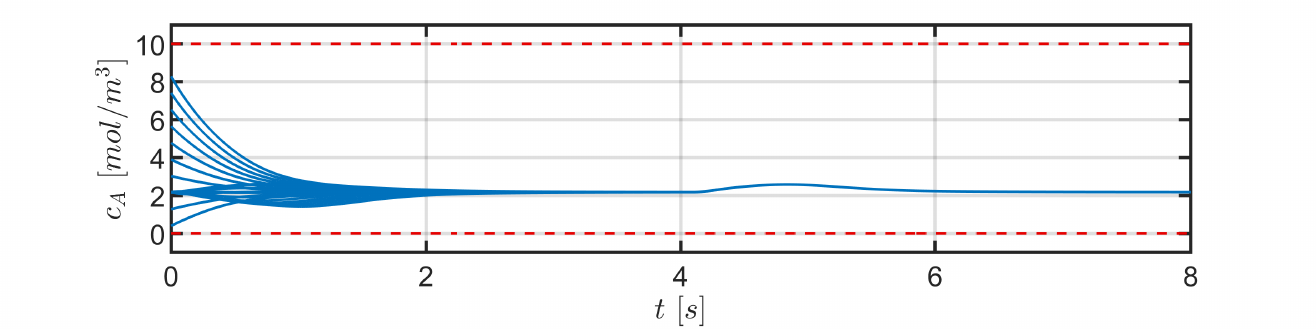}}\\
	\subfigure{\includegraphics[height = 0.17\textheight]{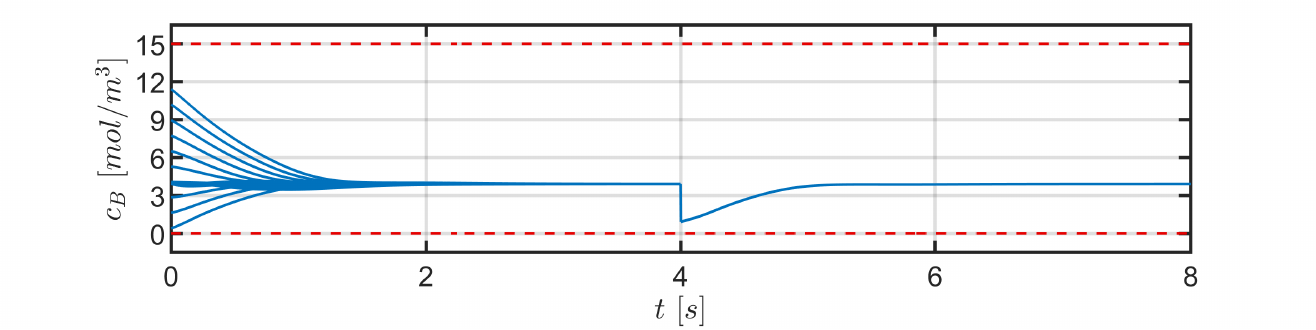}}\\
	\subfigure{\includegraphics[height = 0.17\textheight]{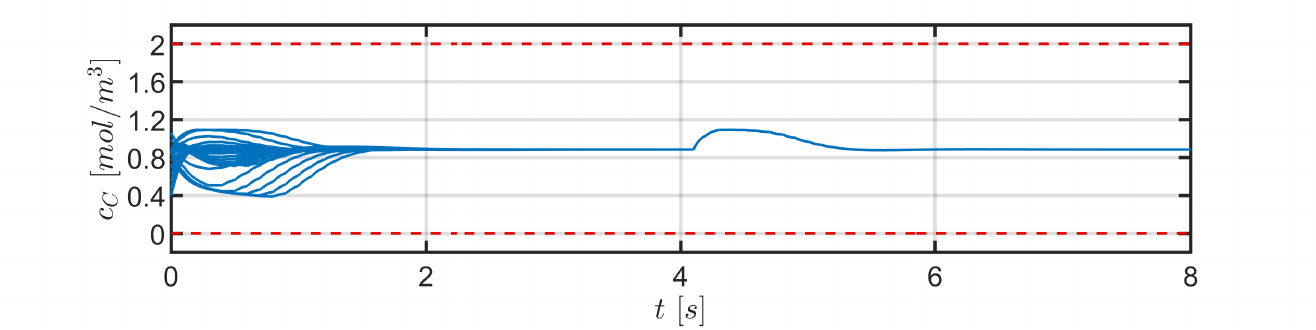}}\\
	\subfigure{\includegraphics[height = 0.17\textheight]{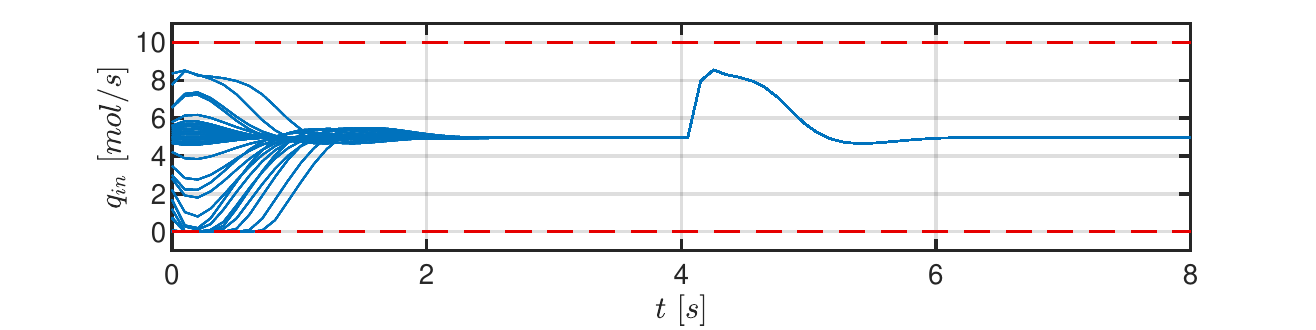}}
	\caption{Control performance of the NN-based controller with various initial conditions.}
	\label{fig:results}
	\vspace{-10pt}
\end{figure}
Finally, we compare and test the applicability of the NN-based controller. A large scale test scenario was prepared, which involved $600$ simulations. Each simulation started from a different initial condition. Thus we can easily observe the performance of the NN-based controller. Furthermore, we introduce an artificial disturbance to the controller variable, to present that the NN-based controller can effectively cope also with disturbances. We especially pointed the reader to the adherence of the bounds on the process and manipulated variables. Control scenarios can be seen in the Figure~\ref{fig:results}, but only a subset of those $600$ simulations are shown, to make the figure readable. We further point out that all simulations are performed using the full-fidelity non-linear model presented in~\eqref{eq:cstr_nonlin}.

Even though the performance of the NN-controller is, in the term of the simulations satisfactory, we further evaluated a quality criterion of the following form
\begin{equation}
    J = \sum_{t=0}^{t_{sim}}\norm{x(t)-x_s}_2^2,
\end{equation}
which indicates how far the actual measurement of process variables was from the desired steady-state value. For each $600$ simulations, we evaluated the $J$ value and compared it to the value of the criterion from the model predictive control performance. The worst decrease in the suboptimality was $2.14\%$.

\section{Conclusions}
The paper discussed the design of suboptimal control law in the form of a neural network. The main advantage of this controller is its explicit form, which was constructed for a large prediction horizon. The neural network was constructed based on data obtained from the optimal solution to the full-fidelity model predictive controller. Applicability of the suboptimal controller was tested on a large scale simulation case study involving the stabilization of multi-component chemical reaction. Simulations results show that in $94.5\%$ of cases, the NN-based explicit controller performed with the less than $1\%$ drop of optimality.

\section*{Acknowledgments} 
Authors gratefully acknowledge the contribution of the Scientific Grant Agency of the Slovak Republic under the grants 1/0585/19. This work was supported by the funding of Slovak Ministry of Education, Science, Research and Sport under the project STU as the Leader of Digital Coalition 002STU-2-1/2018.
M. Klaučo would like to thank for the financial contribution from the STU in Bratislava Grant Scheme for Excellent Research Teams.

\bibliographystyle{apastyle}
\bibliography{bibfile,klauco_ref}
\end{document}

%% file: tikz_pck.tex
\usepackage{etex}
\usepackage{tikz}
\usepackage{pgfkeys}
\usepackage{calc}
\usepackage{pgfplots}
\usetikzlibrary{arrows}
\usepgflibrary{arrows}
\usetikzlibrary{positioning}
\usetikzlibrary{intersections}
\pgfplotsset{compat=newest}
\usetikzlibrary{shapes.geometric}

\makeatletter
\newcommand{\gettikzxy}[3]{%
  \tikz@scan@one@point\pgfutil@firstofone#1\relax
  \edef#2{\the\pgf@x}%
  \edef#3{\the\pgf@y}%
}
\makeatother

\tikzset{
	node style main block/.style={draw, minimum width = 2cm, minimum 
		height=1cm, text height=1.5ex, text depth=.25ex},
	triangle/.style = { regular polygon, regular polygon sides=3, shape border rotate=270,inner sep=1pt}
}

\definecolor{dark_red}{rgb}{0.5, 0.0, 0.0}
\definecolor{dark_blue}{rgb}{0.0, 0.0, 0.6}
\definecolor{light_blue}{rgb}{0.5, 0.5, 1.0}
\definecolor{light_orange}{rgb}{1,0.7490,0.5020}
\definecolor{dark_orange}{rgb}{1,0.5020,0}
\definecolor{light_red}{rgb}{1.0, 0.5, 0.5}
\definecolor{light_green}{rgb}{0.5, 1.0, 0.5}
\definecolor{dark_green}{rgb}{0.0, 0.6, 0.0}
\definecolor{ddark_green}{rgb}{0.0, 0.35, 0.0}
\definecolor{ddark_blue}{rgb}{0.0, 0.0, 0.35}
\definecolor{seahorse}{rgb}{0.839, 0.839, 0.937}
\definecolor{ddark_seahorse}{rgb}{0.739, 0.739, 0.837}
\definecolor{dark_seahorse}{rgb}{0.757, 0.757, 0.909}
\definecolor{g_o_1}{rgb}{0.9725,0.7961,0.6784}
\definecolor{g_o_2}{rgb}{0.9569,0.6902,0.5176}
\definecolor{g_o_3}{rgb}{0.9294,0.4902,0.1922}
\definecolor{g_b_1}{rgb}{0.7412,0.8431,0.9333}
\definecolor{g_b_2}{rgb}{0.6078,0.7608,0.9020}
\definecolor{g_b_3}{rgb}{0.3569,0.6078,0.8353}
\definecolor{excel_good_fill}{rgb}{0.7765,0.9373,0.8078}
\definecolor{excel_good_text}{rgb}{0,0.3804,0}
\definecolor{opacity_text}{rgb}{0.4588,0.4588,0.4824}